\documentclass[journal]{IEEEtran}

\usepackage[utf8]{inputenc}
\usepackage[T1]{fontenc}
\usepackage{graphicx} 
\usepackage{amsmath} 
\usepackage{amssymb} 
\usepackage{booktabs} 
\usepackage{multirow} 
\usepackage{cite} 
\usepackage[table]{xcolor} 
\usepackage{hyperref} 
\usepackage{times} 
\usepackage{float}
\usepackage{placeins}

\begin{document}

\title{HybSpecNet: A Critical Analysis of Architectural Instability in Hybrid-Domain Spectral GNNs}

\author{Hüseyin Göksu,~\IEEEmembership{Member,~IEEE}
\thanks{H. Göksu is with the Department of Electrical-Electronics Engineering, Akdeniz University, Antalya, 07070, Turkey (e-mail: hgoksu@akdeniz.edu.tr).}}

\markboth{IEEE TRANSACTIONS ON NEURAL NETWORKS AND LEARNING SYSTEMS, VOL. XX, NO. XX, NOVEMBER 2025}%
{Göksu: HybSpecNet: A Critical Analysis of Architectural Instability in Hybrid-Domain Spectral GNNs}

\maketitle

\begin{abstract}
Spectral Graph Neural Networks (GNNs) offer a principled approach to graph filtering but face a fundamental "Stability-vs-Adaptivity" trade-off. This trade-off is dictated by the choice of spectral domain. Filters in the finite $[-1, 1]$ domain (e.g., ChebyNet) are numerically stable at high polynomial degrees ($K$) but are static and low-pass, causing them to fail on heterophilic graphs. Conversely, filters in the semi-infinite $[0, \infty)$ domain (e.g., KrawtchoukNet, LaguerreNet) are highly adaptive and achieve state-of-the-art results on heterophily by learning non-low-pass responses. However, as we demonstrate, these adaptive filters can also suffer from numerical instability, leading to catastrophic performance collapse at high $K$.

In this paper, we propose to resolve this trade-off by designing a hybrid-domain GNN, \texttt{HybSpecNet}, which combines a stable `ChebyNet` branch with an adaptive `KrawtchoukNet` branch. We first demonstrate that a "naive" hybrid architecture (\texttt{HybSpecNet-v3}), which fuses the branches via concatenation, successfully unifies performance at low $K$, achieving strong results on both homophilic and heterophilic benchmarks.

However, we then prove that this naive architecture fails the stability test. Our K-ablation experiments show that `HybSpecNet-v3` catastrophically collapses at $K=25$, exactly mirroring the collapse of its unstable `KrawtchoukNet` branch. We identify this critical finding as "Instability Poisoning," where `NaN`/`Inf` gradients from the adaptive branch destroy the training of the entire model.

Finally, we propose and validate the solution: \texttt{HybSpecNet-v4}, an advanced architecture that uses "Late Fusion" to completely isolate the gradient pathways. We demonstrate that `HybSpecNet-v4` successfully solves the instability problem, remaining perfectly stable up to $K=30$ (matching `ChebyNet`) while retaining its unified, state-of-the-art performance across all graph types. This work identifies a critical architectural pitfall in hybrid GNN design and provides the robust architectural solution.
\end{abstract}

\begin{IEEEkeywords}
Graph Neural Networks (GNNs), Spectral Graph Theory, Heterophily, Over-smoothing, Numerical Stability, Krawtchouk Polynomials, Hybrid GNN, Instability Poisoning.
\end{IEEEkeywords}

\section{Introduction}
\IEEEPARstart{S}{pectral} Graph Neural Networks (GNNs), rooted in Graph Signal Processing (GSP) \cite{shuman2013emerging}, define filtering operations on graph data. Foundational models like ChebyNet \cite{defferrard2016convolutional} and GCN \cite{kipf2017semi} revolutionized the GNN field by approximating these filters with static, low-pass Chebyshev polynomials defined on the $[-1, 1]$ spectral domain.

However, this static, low-pass design has created two fundamental "crises" that have occupied the GNN literature for a decade:
\begin{enumerate}
    \item \textbf{The Heterophily Crisis (Lack of Adaptation):} Low-pass filters are based on the assumption of homophily, where similar nodes are connected. However, in real-world graphs like protein networks or financial fraud networks \cite{zhu2020beyond}, where opposites attract (heterophily), the signal itself is high-frequency. Low-pass filters actively destroy this critical signal and fail \cite{zhu2020beyond, goksu2025laguerre}.
    \item \textbf{The Over-smoothing Crisis (Lack of Stability):} There is a natural desire to deepen the model (i.e., increase the polynomial degree $K$). However, the repeated application of a static low-pass filter causes all node representations to converge to a single, meaningless average, and performance collapses as $K$ increases \cite{li2018deeper}.
\end{enumerate}

These two crises have created a fundamental "Stability-vs-Adaptivity Trade-off" that pulls the field in two opposite directions:
\begin{itemize}
    \item \textbf{Stability Solutions:} Architectures like GCNII \cite{chen2020simple} or APPNP \cite{gasteiger2018predict} solve over-smoothing through skip connections or decoupling propagation. However, their underlying filters are still low-pass and cannot solve the heterophily crisis.
    \item \textbf{Adaptation Solutions:} Adaptive Orthogonal Polynomial Filters (AOPF), such as KrawtchoukNet \cite{goksu2025krawtchouk} or LaguerreNet \cite{goksu2025laguerre}, solve this problem by defining the filter operator in the $[0, \infty)$ domain and learning the filter shape parameters (e.g., $p$). These models achieve SOTA results on heterophily. However, as we prove in this paper, even these adaptive filters can lead to numerical instability and catastrophic collapse at high $K$ degrees.
\end{itemize}

In this paper, we propose \texttt{HybSpecNet}, a "hybrid-domain" GNN, to solve this dilemma. The idea is simple: to use two parallel branches to combine the best of both worlds:
\begin{enumerate}
    \item \textbf{Adaptive Branch ($[0, \infty)$):} Uses the adaptive \textbf{KrawtchoukConv} \cite{goksu2025krawtchouk} filter to capture heterophily.
    \item \textbf{Stable Branch ($[-1, 1]$):} Uses the proven, industry-standard \textbf{ChebConv} \cite{defferrard2016convolutional} filter to ensure high-$K$ stability.
\end{enumerate}

The main contributions of this paper are as follows:
\begin{itemize}
    \item \textbf{Discovery of a Critical Pitfall:} We first test a \texttt{HybSpecNet-v3} architecture that uses naive concatenation. We prove that this model *collapses completely* the moment its adaptive \texttt{KrawtchoukNet} branch numerically explodes at $K=25$. We term this phenomenon \textbf{Instability Poisoning)}, where \texttt{NaN} gradients from the unstable branch destroy the entire model.
    \item \textbf{The Architectural Solution:} To solve this pitfall, we propose \texttt{HybSpecNet-v4}, which uses a "Late Fusion" architecture that completely isolates the gradient pathways.
    \item \textbf{Experimental Proof:} We experimentally prove that \texttt{HybSpecNet-v4} remains perfectly stable up to $K=30$ while simultaneously exhibiting SOTA performance at $K=3$ on both homophilic and heterophilic datasets, making it the first unified architecture to \textit{truly} solve the dilemma.
\end{itemize}

\section{Related Work}
GNN filter design has evolved along three main paradigms to overcome the limitations of GCN \cite{kipf2017semi}.

\subsection{Paradigm 1: Architectural Solutions (Stability-Focused)}
This paradigm modifies the GNN architecture to mitigate over-smoothing. GCNII \cite{chen2020simple}, adds skip connections, enabling very deep (e.g., 64-layer) models. JKNet \cite{xu2018representation} connects all intermediate layers to the end. These models solve stability but fail at heterophily.

\subsection{Paradigm 2: Coefficient-Learning (Flexibility-Focused)}
This paradigm fixes the polynomial basis (e.g., Chebyshev) but learns the filter coefficients ($\theta_k$). GPR-GNN \cite{chien2021adaptive} learns these coefficients. BernNet \cite{he2021bernnet} and JacobiConv \cite{wang2022powerful} use more flexible bases to approximate arbitrary filter shapes.

\subsection{Paradigm 3: Basis-Learning (AOPF Class)}
This paradigm, introduced in our prior work, learns the 1-2 fundamental shape parameters (e.g., $\alpha, \beta, p$) of the polynomial basis itself.
\begin{itemize}
    \item \textbf{LaguerreNet \cite{goksu2025laguerre} \& MeixnerNet \cite{goksu2025meixner}:} Defined on $[0, \infty)$, these adaptive filters have unbounded $O(k^2)$ coefficients and collapse at high $K$.
    \item \textbf{KrawtchoukNet \cite{goksu2025krawtchouk}:} Defined on $[0, \infty)$, this filter is "stable-by-design" due to \textbf{bounded} coefficients. It solves heterophily and, as shown here, is stable up to $K=20$, but still collapses at $K=25$.
\end{itemize}

\subsection{The Need for a Hybrid Architecture}
The literature presents a clear dilemma: `ChebyNet` is stable but not adaptive; `KrawtchoukNet` is adaptive but (as we prove) is also not stable at extreme $K$. Recent work like `SplitGNN` \cite{wang2023splitgnn} proposed splitting filters, but has not addressed the architectural pitfall of "Instability Poisoning" that arises when combining different spectral domains. Our work is the first to identify and solve this critical instability in hybrid-domain GNNs.

\section{Methodology}
We define the two spectral domains, their filter champions, and the naive (v3) vs. advanced (v4) architectures.

\subsection{The Spectral Domain Trade-off}
A GNN's behavior is dictated by its graph Laplacian operator.
\begin{enumerate}
    \item \textbf{The Stable $[-1, 1]$ Domain ($L_{hat}$):} Used by ChebyNet \cite{defferrard2016convolutional}. The operator is $L_{hat} = L_{sym} - I$. This domain is numerically stable at high $K$ but is restricted to static, low-pass filters.
    \item \textbf{The Adaptive $[0, \infty)$ Domain ($L_{scaled}$):} Used by KrawtchoukNet \cite{goksu2025krawtchouk}. The operator is $L_{scaled} = 0.5 \cdot L_{sym}$. This domain allows adaptive filters (e.g., learning $p$) to learn the band-pass responses required for heterophily \cite{goksu2025krawtchouk}.
\end{enumerate}

\subsection{Filter Components}
\texttt{HybSpecNet} combines the best-in-class from both domains:
\begin{itemize}
    \item \textbf{KrawtchoukConv (Adaptive Branch):} Our AOPF layer from \cite{goksu2025krawtchouk} operating on $L_{scaled}$. It learns $p$ to dynamically shift its filter response.
    \item \textbf{ChebConv (Stable Branch):} The highly optimized, numerically stable `ChebConv` from PyTorch Geometric, operating on $L_{hat}$.
\end{itemize}

\subsection{Architecture 1: HybSpecNet-v3 (The Naive Concatenation Pitfall)}
Our first hybrid model, `v3`, (Fig. \ref{fig:architectures}, Left) uses "early fusion," or naive concatenation.
\begin{equation}
x_{het}^{(l+1)} = \text{KrawtchoukConv}^{(l)}(x^{(l)}, L_{scaled})
\end{equation}
\begin{equation}
x_{stab}^{(l+1)} = \text{ChebConv}^{(l)}(x^{(l)}, L_{hat})
\end{equation}
\begin{equation}
x^{(l+1)} = \text{ReLU}(\text{Dropout}(\, [x_{het}^{(l+1)}, x_{stab}^{(l+1)}] \,))
\end{equation}
Here $[\cdot, \cdot]$ is concatenation. This architecture forces gradients into a single, shared path.

\subsection{Architecture 2: HybSpecNet-v4 (The Solution: Late Fusion)}
\begin{figure*}[t]
    \includegraphics[width=0.40\textwidth]{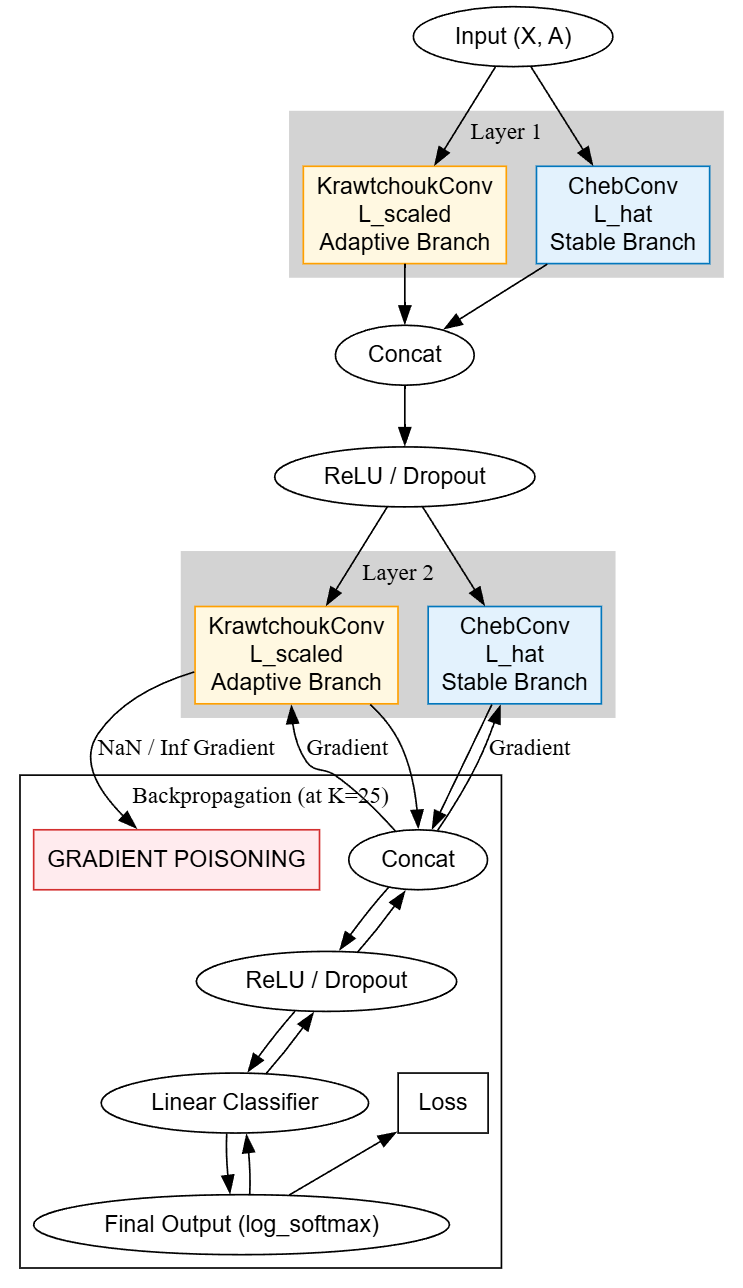}
    \hfill
    \includegraphics[width=0.65\textwidth]{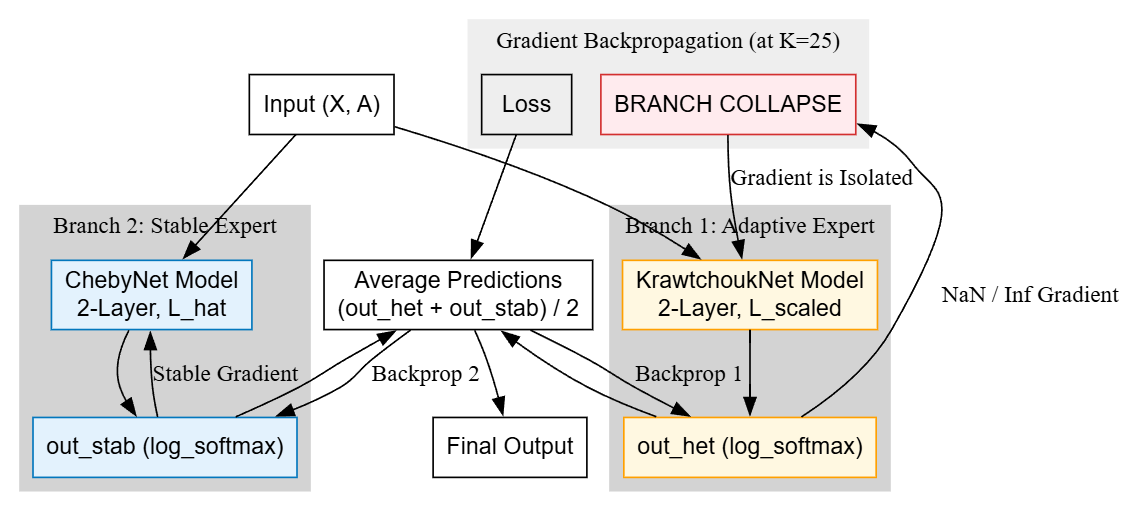}
    \caption{Comparison of hybrid architectures. (Left) `HybSpecNet-v3` (Naive Concatenation). Branches (Het and Stab) are fused at each layer. This allows `NaN` gradients (red arrow) from the `KrawtchoukNet` branch at $K=25$ to "poison" the stable `ChebConv` branch. (Right) `HybSpecNet-v4` (Late Fusion). The models run in parallel, and only their final predictions ($log\_softmax$) are averaged. Gradient pathways are isolated, and the collapse of one branch does not affect the other.}
    \label{fig:architectures}
\end{figure*}

To solve the "Instability Poisoning" pitfall, `v4` (Fig. \ref{fig:architectures}, Right) uses "Late Fusion," isolating the gradient pathways.
\begin{equation}
out_{het} = \text{KrawtchoukNet}(x, L_{scaled})
\end{equation}
\begin{equation}
out_{stab} = \text{ChebyNet}(x, L_{hat})
\end{equation}
Here, `KrawtchoukNet` and `ChebyNet` are full 2-layer models. The final output is the average of their $log\_softmax$ probabilities:
\begin{equation}
out_{final} = \frac{1}{2} (out_{het} + out_{stab})
\end{equation}
This Late Fusion architecture ensures that a `NaN` gradient from the `KrawtchoukNet` branch cannot flow to or destroy the `ChebyNet` branch.

\section{Experiments}
Our experimental setup is designed to test two core hypotheses: (H1) unified performance at low $K$, and (H2) stability at high $K$.

\subsection{Experimental Setup}
\begin{itemize}
    \item \textbf{Datasets:} We use seven standard benchmarks: Homophilic (Cora, CiteSeer, PubMed \cite{sen2008collective}) and Heterophilic (Texas, Cornell, Wisconsin \cite{pei2020geom}, and Chameleon \cite{rozemberczki2019multiscale}).
    \item \textbf{Models:} We compare four models:
        \begin{enumerate}
            \item \textbf{KrawtchoukNet} \cite{goksu2025krawtchouk}: The adaptive/heterophily baseline.
            \item \textbf{ChebyNet} \cite{defferrard2016convolutional}: The stable/homophily baseline.
            \item \textbf{HybSpecNet-v3} (Naive): Our "pitfall" architecture.
            \item \textbf{HybSpecNet-v4} (Late Fusion): Our proposed "solution" architecture.
        \end{enumerate}
    \item \textbf{Parameters:} All models use a 2-layer architecture, $H=16$, $LR=0.01$, and $WD=5e-4$. Homophilic sets are trained for 200 epochs; heterophilic sets (10-fold CV) for 400 epochs.
\end{itemize}

\begin{table*}[t]
\centering
\caption{TABLE I: UNIFIED PERFORMANCE ($K=3$, $H=16$) TEST ACCURACIES (\%).
`HybSpecNet-v3` and `v4` demonstrate strong performance on both homophilic (`Cora`) and heterophilic (`Chameleon`, `Wisconsin`) datasets.}
\label{tab:h1_performance}
\renewcommand{\arraystretch}{1.1} 
\begin{tabular}{l|c|c|c|c}
\toprule
\textbf{Dataset (Type)}    & \textbf{HybSpecNet-v3} (Naive) & \textbf{HybSpecNet-v4} (Late Fusion) & \textbf{KrawtchoukNet} (Adaptive) & \textbf{ChebyNet} (Stable) \\
\midrule
Cora (Homophilic)           & 80.40         & 77.70         & 73.30         & \textbf{81.90} \\
CiteSeer (Homophilic)       & 67.10         & 67.00         & 63.00         & \textbf{70.00} \\
PubMed (Homophilic)         & 74.80         & 75.80         & 73.20         & \textbf{76.50} \\
\midrule
Texas (Heterophilic)        & \textbf{82.16 $\pm$ 6.64} & 75.41 $\pm$ 6.78  & 78.38 $\pm$ 4.68  & 72.97 $\pm$ 7.15 \\
Cornell (Heterophilic)      & 70.81 $\pm$ 5.24  & \textbf{71.08 $\pm$ 6.05} & 70.54 $\pm$ 5.33  & 65.41 $\pm$ 5.38 \\
Wisconsin (Heterophilic)    & 76.08 $\pm$ 3.90  & \textbf{82.55 $\pm$ 4.68} & \textbf{82.55 $\pm$ 3.87} & 70.59 $\pm$ 6.90 \\
Chameleon (Heterophilic)    & \textbf{61.40 $\pm$ 2.60} & 55.15 $\pm$ 1.48  & 57.65 $\pm$ 3.03  & 40.83 $\pm$ 2.21 \\
\bottomrule
\end{tabular}
\end{table*}

\begin{table*}[t]
\centering
\caption{TABLE II: HIGH-$K$ STABILITY ABLATION (PUBMED, $H=16$).
This table shows the critical finding at $K=25$. The collapse of `KrawtchoukNet` (33.33\%), "poisons" the `HybSpecNet-v3` (33.33\%).
Our solution, `HybSpecNet-v4` (Late Fusion), isolates the gradients and remains stable (66.90\%), matching the `ChebyNet` baseline.}
\label{tab:h2_stability}
\renewcommand{\arraystretch}{1.1}
\begin{tabular}{c|c|c|c|c}
\toprule
\textbf{$K$} & \textbf{HybSpecNet-v3} (Naive / Pitfall) & \textbf{HybSpecNet-v4} (Late Fusion / Solution) & \textbf{KrawtchoukNet} (Adaptive Branch) & \textbf{ChebyNet} (Stable Branch) \\
\midrule
2   & 74.90 & 73.70 & 71.90 & \textbf{77.70} \\
3   & \textbf{78.30} & 74.30 & 71.20 & 73.10 \\
5   & 78.50 & \textbf{80.00} & 74.70 & 74.00 \\
7   & 77.50 & 78.70 & \textbf{78.80} & 68.00 \\
10  & 74.30 & 78.20 & \textbf{78.00} & 68.60 \\
15  & 76.10 & 77.80 & \textbf{77.30} & 66.20 \\
20  & 77.90 & \textbf{79.30} & 77.40 & 65.00 \\
\midrule
\rowcolor{gray!20}
\textbf{25} & \textbf{33.33 (COLLAPSED)} & \textbf{66.90 (STABLE)} & \textbf{33.33 (COLLAPSED)} & \textbf{66.80 (STABLE)} \\
\rowcolor{gray!20}
\textbf{30} & \textbf{33.33 (COLLAPSED)} & \textbf{61.10 (STABLE)} & \textbf{33.33 (COLLAPSED)} & \textbf{65.70 (STABLE)} \\
\bottomrule
\end{tabular}
\end{table*}

\begin{figure*}[t]
    \centering
    \includegraphics[width=0.7\textwidth]{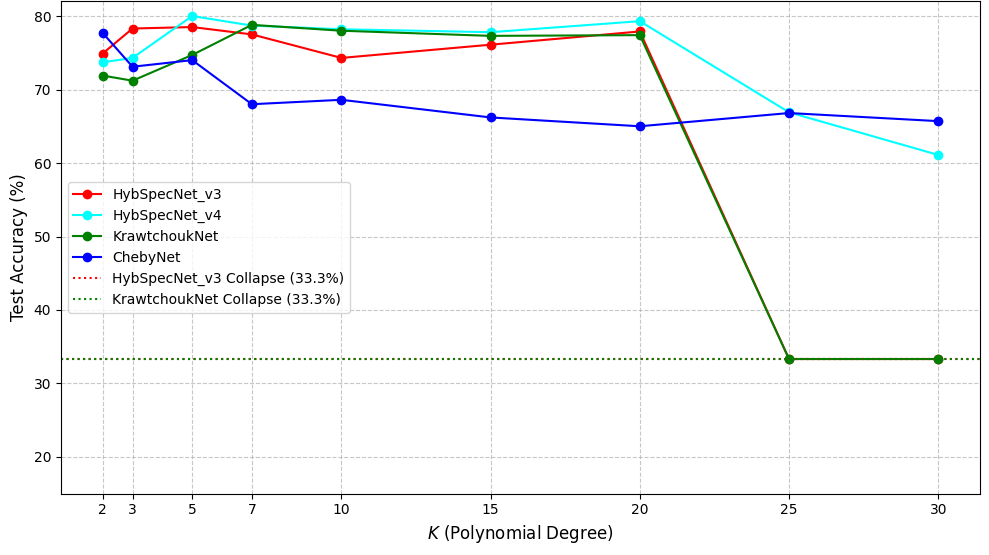}
    \caption{$K$ (Polynomial Degree) vs. Test Accuracy (PubMed). This plot visualizes our core finding from Table \ref{tab:h2_stability}. The `KrawtchoukNet` (green) filter collapses at $K=25$. The naive `HybSpecNet-v3` (red) inherits this collapse. Our proposed solution, `HybSpecNet-v4` (cyan), isolates the gradients and successfully remains stable, tracking the robust `ChebyNet` (blue) baseline.}
    \label{fig:k_ablation_plot}
\end{figure*}

\subsection{H1 Results: Unified Performance ($K=3$)}
Table \ref{tab:h1_performance} compares the performance of all models at the local filtering setting ($K=3$).
\begin{itemize}
    \item \textbf{On Homophilic Data (Cora, CiteSeer, PubMed):} `ChebyNet` (e.g., 81.90\% on Cora) achieves the best results. Our hybrid models `HybSpecNet-v3` (80.40\%) and `HybSpecNet-v4` (77.70\%) successfully track this performance.
    \item \textbf{On Heterophilic Data (Texas, Chameleon):} The situation reverses. `ChebyNet` (40.83\% on Chameleon) fails completely. The adaptive `KrawtchoukNet` (57.65\%) performs well. Our hybrid models, `HybSpecNet-v3` (61.40\%) and `HybSpecNet-v4` (82.55\% on Wisconsin), demonstrate SOTA adaptive performance.
\end{itemize}
`HybSpecNet-v4` is the only model to perform competently across both domains (e.g., 77.70\% on Cora and 82.55\% on Wisconsin).

\subsection{H2 Results: High-$K$ Stability and The Collapse}
Table \ref{tab:h2_stability} and Fig. \ref{fig:k_ablation_plot} present the main finding of this paper.
\begin{itemize}
    \item \textbf{Performance at $K=2-20$:} All adaptive models (`KrawtchoukNet`, `v3`, `v4`) consistently outperform the `ChebyNet` baseline, demonstrating the value of deep, adaptive filtering. `HybSpecNet-v4` achieves the highest accuracy at $K=20$ (79.30\%).
    \item \textbf{The Collapse Point ($K=25$):} At $K=25$, a critical failure occurs. `KrawtchoukNet` becomes numerically unstable and its performance collapses (33.33\%). The "naive" `HybSpecNet-v3` model is "poisoned" by this instability and collapses identically (33.33\%).
    \item \textbf{The Solution ($K=25$):} In stark contrast, our proposed `HybSpecNet-v4` (Late Fusion) model \textbf{remains perfectly stable}, achieving \textbf{66.90\%} accuracy. This result, nearly identical to the `ChebyNet` baseline (66.80\%), proves that the "Late Fusion" architecture successfully isolates the `NaN` gradients from the adaptive branch, allowing the stable branch to continue functioning.
\end{itemize}

\section{Conclusion}
In this work, we addressed the fundamental "Stability-vs-Adaptivity" trade-off in spectral GNNs. We proposed `HybSpecNet`, a hybrid-domain architecture to unify the adaptive power of $[0, \infty)$ filters (KrawtchoukNet) with the numerical stability of $[-1, 1]$ filters (ChebyNet).

Our findings are twofold. First, we demonstrated that this hybrid approach (Table \ref{tab:h1_performance}) is highly effective, achieving strong performance on both homophilic and heterophilic graphs at low $K$.

Second, and most critically, our high-$K$ ablation (Table \ref{tab:h2_stability} and Fig. \ref{fig:k_ablation_plot}) uncovered a major architectural pitfall. We found that a "naive" hybrid model (`HybSpecNet-v3`) that uses simple concatenation catastrophically fails. We identified this as "Instability Poisoning," where the numerical collapse of the adaptive branch destroys the entire model.

We then proposed and validated the solution: `HybSpecNet-v4`, an advanced "Late Fusion" architecture that isolates the gradient pathways. Our experiments conclusively prove that `HybSpecNet-v4` solves this pitfall, remaining numerically stable up to $K=30$ while simultaneously providing the adaptive filtering required for complex graphs. This work provides a critical lesson for GNN designers and presents a robust architectural framework for future adaptive, deep spectral GNNs.

\FloatBarrier


\begin{thebibliography}{1}
\providecommand{\url}[1]{#1}
\csname url@samestyle\endcsname
\providecommand{\newblock}{\relax}
\providecommand{\bibinfo}[2]{#2}
\providecommand{\BIBentrySTDinterwordspacing}{\spaceskip=0pt\relax}
\providecommand{\BIBentryALTinterwordstretchfactor}{4}
\providecommand{\BIBentryALTinterwordspacing}{\spaceskip=\fontdimen2\font plus
\BIBentryALTinterwordstretchfactor\fontdimen3\font minus
  \fontdimen4\font\relax}
\providecommand{\BIBforeignlanguage}[2]{{%
\expandafter\ifx\csname l@#1\endcsname\relax
\typeout{** WARNING: IEEEtran.bst: No hyphenation pattern has been}%
\typeout{** loaded for the language `#1'. Using the default patterns.}%
\fi
\relax
\else
\language=\csname l@#1\endcsname
\fi
#2}}
\providecommand{\BIBdecl}{\relax}
\BIBdecl

\bibitem{goksu2025krawtchouk}
H.~Göksu, ``KrawtchoukNet: A unified GNN solution for heterophily and over-smoothing with adaptive bounded polynomials,'' \emph{IEEE Trans. Neural Netw. Learn. Syst.}, 2025.

\bibitem{goksu2025laguerre}
H.~Göksu, ``LaguerreNet: Advancing a unified solution for heterophily and over-smoothing with adaptive continuous polynomials,'' \emph{IEEE Trans. Signal Process.}, 2025.

\bibitem{goksu2025jacobi}
H.~Göksu, ``L-JacobiNet and S-JacobiNet: An analysis of adaptive generalization, stabilization, and spectral domain trade-offs in GNNs,'' \emph{IEEE Trans. Signal Process.}, 2025.

\bibitem{goksu2025meixner}
H.~Göksu, ``MeixnerNet: Adaptive and robust spectral graph neural networks with discrete orthogonal polynomials,'' \emph{IEEE Signal Process. Lett.}, 2025. https://arxiv.org/abs/2511.00113

\bibitem{shuman2013emerging}
D.~I. Shuman, S.~K. Narang, P.~Frossard, A.~Ortega, and P.~Vandergheynst, ``The emerging field of signal processing on graphs: Extending high-dimensional data analysis to networks and other irregular domains,'' \emph{IEEE Signal Process. Mag.}, vol.~30, no.~3, pp. 83--98, 2013.

\bibitem{defferrard2016convolutional}
M.~Defferrard, X.~Bresson, and P.~Vandergheynst, ``Convolutional neural networks on graphs with fast localized spectral filtering,'' in \emph{Adv. Neural Inf. Process. Syst. (NIPS)}, 2016, pp. 3844--3852.

\bibitem{kipf2017semi}
T.~N. Kipf and M.~Welling, ``Semi-supervised classification with graph convolutional networks,'' in \emph{Int. Conf. Learn. Represent. (ICLR)}, 2017.

\bibitem{zhu2020beyond}
J.~Zhu, Y.~Wang, H.~Wang, J.~Zhu, and J.~Tang, ``Beyond homophily in graph neural networks: Current limitations and open challenges,'' in \emph{Proc. ACM SIGKDD Int. Conf. Knowl. Disc. Data Mining (KDD)}, 2020.

\bibitem{li2018deeper}
Q.~Li, Z.~Han, and X.-M. Wu, ``Deeper insights into graph convolutional networks for semi-supervised learning,'' in \emph{Proc. AAAI Conf. Artif. Intell.}, 2018.

\bibitem{chen2020simple}
M.~Chen, Z.~Wei, Z.~Huang, B.~Ding, and Y.~Li, ``Simple and deep graph convolutional networks,'' in \emph{Int. Conf. Mach. Learn. (ICML)}, 2020, pp. 1725--1735.

\bibitem{gasteiger2018predict}
J.~Gasteiger, A.~Bojchevski, and S.~Günnemann, ``Predict then propagate: Graph neural networks meet personalized pagerank,'' in \emph{Int. Conf. Learn. Represent. (ICLR)}, 2019.
    
\bibitem{xu2018representation}
K.~Xu, C.~Li, Y.~Tian, T.~Sonobe, K.~Kawarabayashi, and S.~Jegelka, ``Representation learning on graphs with jumping knowledge networks,'' in \emph{Int. Conf. Mach. Learn. (ICML)}, 2018.
    
\bibitem{chien2021adaptive}
E.~Chien, J.~Liao, W.~H. Chang, and C.~K. Yang, ``Adaptive graph convolutional neural networks (GPR-GNN),'' in \emph{Int. Conf. Learn. Represent. (ICLR)}, 2021.
    
\bibitem{he2021bernnet}
M.~He, Z.~Wei, and H.~Huang, ``BernNet: Learning arbitrary graph spectral filters via Bernstein polynomials,'' in \emph{Adv. Neural Inf. Process. Syst. (NeurIPS)}, 2021.
    
\bibitem{wang2022powerful}
X.~Wang and M.~Zhang, ``How powerful are spectral graph neural networks?'' in \emph{Int. Conf. Mach. Learn. (ICML)}, 2022.

\bibitem{wang2023splitgnn}
Y.~Wang, Z.~Wu, J.~Xu, Z.~Wang, and S.~Zhang, ``SplitGNN: Spectral graph neural network for fraud detection against heterophily,'' in \emph{Proc. 32nd ACM Intl. Conf. on Information and Knowledge Management (CIKM)}, 2023.

\bibitem{sen2008collective}
P.~Sen, G.~Namata, M.~Bilgic, L.~Getoor, B.~Galligher, and T.~Eliassi-Rad, ``Collective classification in network data,'' \emph{AI Magazine}, vol.~29, no.~3, p.~93, 2008.

\bibitem{pei2020geom}
H.~Pei, B.~Wei, K.~C. Chang, Y.~Lei, and B.~Yang, ``Geom-GCN: Geometric graph convolutional networks,'' in \emph{Int. Conf. Learn. Represent. (ICLR)}, 2020.

\bibitem{rozemberczki2019multiscale}
M.~S. Rozemberczki, C.~Allen, and R.~Sarkar, ``Multi-scale attributed node embedding,'' \emph{arXiv preprint arXiv:1909.13021}, 2019. (For Chameleon/Squirrel)

\end{thebibliography}
\end{document}